\documentclass[10pt, a4paper]{article}
\usepackage{lrec}
\usepackage{graphicx}
\usepackage{tabularx}
\usepackage{booktabs}
\usepackage{soul}
\usepackage{amssymb}
\usepackage{amsmath}
\usepackage{footmisc}

\usepackage{epstopdf}
\usepackage[latin1]{inputenc}

\usepackage{xcolor}
\definecolor{darkblue}{rgb}{0.0,0.0,0.35}
\usepackage{hyperref}
\hypersetup{
    colorlinks = True,
    allcolors = darkblue
}
\usepackage{xstring}

\usepackage{tikz}
\usepackage{pgfplots}
\usepackage{pgfplotstable}
\usepgfplotslibrary{fillbetween}
\usepgfplotslibrary{colorbrewer}
\pgfplotsset{compat=1.14, colormap/Blues, every axis/.append style={label style={font=\footnotesize}, tick label style={font=\footnotesize}}}

\DeclareMathOperator*{\argmax}{arg\,max}

\newcommand{\factor}{\text{factor}}

\newcommand\blfootnote[1]{%
  \begingroup
  \renewcommand\thefootnote{}\footnote{$^\dagger$#1}%
  \addtocounter{footnote}{-1}%
  \endgroup
}

\title{A Robust Self-Learning Method for Fully Unsupervised Cross-Lingual Mappings of Word Embeddings: Making the Method Robustly Reproducible as Well}

\name{Nicolas Garneau$^\dagger$, Mathieu Godbout$^\dagger$, David Beauchemin$^\dagger$, Audrey Durand, Luc Lamontagne}

\address{Universit\'e Laval \\
         2325 Rue de l'Universit\'e, Qu\'ebec \\
         nicolas.garneau@ift.ulaval.ca, mathieu.godbout.3@ulaval.ca, david.beauchemin.5@ulaval.ca\\
         audrey.durand@ift.ulaval.ca, luc.lamontagne@ift.ulaval.ca}

\abstract{
In this paper, we reproduce the experiments of \newcite{artetxe-etal-2018-robust} regarding the robust self-learning method for fully unsupervised cross-lingual mappings of word embeddings. We show that the reproduction of their method is indeed feasible with some minor assumptions. We further investigate the robustness of their model by introducing four new languages that are less similar to English than the ones proposed by the original paper. In order to assess the stability of their model, we also conduct a grid search over sensible hyperparameters. We then propose key recommendations that apply to any research project in order to deliver fully reproducible research. \newline
}

\begin{document}

\maketitleabstract

\section{Introduction}
\blfootnote{Authors contributed equally to this work.}
The cross-lingual mapping of word embeddings is a problem that has been studied more thoroughly with the rise of distributed representations induced from neural network architectures \cite{Mikolov2013DistributedRO}.
The goal of this task is to induce automatically a word-to-word translation dictionary $\mathbf{D} \in \mathbb{R}^{|\mathcal{V}_s| \times |\mathcal{V}_t|}$ where $\mathbf{D}[i,j] = 1$ means that the $i$-th word from the source vocabulary $\mathcal{V}_s$ is a translation of the $j$-th word in the target vocabulary $\mathcal{V}_t$.
The data used to learn the mapping from two different languages is two sets of word embeddings $\mathbf{X}_s$ and $\mathbf{X}_t$ corresponding to the vectors of the source and the target language respectively.
The mapping from one language space to the other is usually done with a projection matrix $\mathbf{W}_t$ which projects the embeddings of the target language in the same space as the source language (or vice versa), i.e. $\mathbf{X}_s \approx \mathbf{X}_t\mathbf{W}_t$.

There are several methods available to achieve such a mapping depending on the resources in hand.
Given a dataset of parallel word-aligned data, supervised mapping-based approaches are amongst the most popular to date  \cite{Mikolov2013ExploitingSA,Dinu2014ImprovingZL,Gouws2015SimpleTB}.
Several unsupervised methods \cite{Yang2018LearningUW,conneau2017word} based on Generative Adversarial Neural Networks (GANs) \cite{Goodfellow2014GenerativeAN} have also been proposed in the case where no seed dictionary is available.

The method that we reproduce is the work of \newcite{artetxe-etal-2018-robust} (referenced as \textit{"the authors"}). It also falls in the unsupervised setting but is based on distances between nearest neighbors to build the initial seed dictionary.
We refer the reader to the survey of \newcite{Ruder2018ASO} for more details, which provides an extensive overview of the different methods for learning cross-lingual mappings between two different word embedding spaces.

On another hand, delivering reproducible research is too often an underestimated concern.
The fast pace of the research community makes the verification of every submitted paper barely possible, especially in a boiling community such as natural language processing.
Fortunately, we see the rise of different challenges\footnote{\url{https://reproducibility-challenge.github.io/iclr_2019/}}\textsuperscript{,}\footnote{\url{https://aaai.org/Conferences/AAAI-19/ws19workshops/\#ws16}}\textsuperscript{,}\footnote{\url{http://rescience.github.io/}} that emphasize the importance of supporting the proposed results with an official code implementation as well as the corresponding dataset.
It has even been mandatory to provide the source code and the detailed procedure to obtain the same results like the ones claimed in the paper in the NeurIPS Reproducibility Challenge\footnote{\url{https://reproducibility-challenge.github.io/neurips2019/}}.

In light of this quest for reproducibility, we hereby propose to reproduce the paper from \newcite{artetxe-etal-2018-robust} in the context of REPROLANG 2020 by also providing the stammering of a methodology for delivering reproducible experiments.
Even though this algorithm relies on a stochastic component, we can reproduce the results issued from this algorithm and further analyze its behavior.
It is therefore important to provide a clean, readable codebase that supports a clear and concise paper.

We begin in \autoref{sec:problem} with the problem statement and \autoref{sec:algorithm} with the presentation of the analyzed algorithm.
We then present what we reproduced from the original paper's results as well as how we obtained our results in \autoref{sec:reproducing}
We provide recommendations regarding the techniques used to obtain these results and their applicability to other research projects in \autoref{sec:recommandations} and close the analysis with an assessment of the algorithm's robustness in \autoref{sec:assess}
We also publish our code online as required \footnote{\url{https://gitlab.com/nicolasgarneau/vecmap/} using this commit: \href{https://gitlab.com/nicolasgarneau/vecmap/commit/b1abbd2635d6174b8b10d75e7ca5a652249b6841}{\texttt{b1abbd26}}}.

\section{Problem Statement}
\label{sec:problem}

Word vectors, often called \textit{word embeddings}, are distributed representations derived from a textual corpus \cite{Mikolov2013DistributedRO,Pennington2014GloveGV}.
The dimension $d$ of these representations often spans from 100 to 1,000.
One key outcome of learning these word representations is that it associates words in a vocabulary $\mathcal{V}$ with a similar meaning (appearing in a similar context) with similar distribution vectors.
This set of representations, often called the embedding matrix $\mathbf{X} \in \mathbb{R}^{|\mathcal{V}| \times d}$ serves as the input for many models in natural language understanding applications such as text classification and machine translation.

The paper on which we conduct our reproducibility experiment tackles the task of word vectors space alignment.
Given two sets of source and target embedding matrices $\mathbf{X}_s \in \mathbb{R}^{|\mathcal{V}_s| \times d}$ and $\mathbf{X}_t \in \mathbb{R}^{|\mathcal{V}_t| \times d}$ induced from two different textual corpora, one tries to find the mappings $\mathbf{W}_s \in \mathbb{R}^{d \times d}$ and $\mathbf{W}_t \in \mathbb{R}^{d \times d}$ such that $\mathbf{X}_s\mathbf{W}_s \approx \mathbf{X}_t\mathbf{W}_t$.
The vocabularies $\mathcal{V}_s$ and $\mathcal{V}_t$ can be of different sizes.
Usually, only a subset of the $n$ most frequent words is used for the alignment.
The work of the authors strictly focuses on the task of \textbf{unsupervised bilingual dictionary induction}, hence aligning the word vector space of a source language with one of a target language without word-aligned data.

Essentially, the original paper's approach tries to find a good initial solution $\mathbf{D}^0$ by aligning word vectors from the source and the target language that have a similar distribution.
Their motivation, referred to in the literature as the isometry assumption, is that monolingual word embedding spaces are approximately isomorphic \cite{vulic-etal-2019-really}.
They demonstrated for example that the vector of the word ``two'' in English has a similar distribution as the word vector ``due'' in Italian and will be different from the distribution of the word ``cane'', also in Italian.
Once the initial dictionary $\mathbf{D}^0$ is induced from the unsupervised procedure, a self-supervised iteration loop is invoked to refine the mapping from the source to the target language.
The whole algorithm is further analyzed in the following section.

\section{The Proposed Algorithm}
\label{sec:algorithm}

In this section, we detail the four different steps of the algorithm proposed by \newcite{artetxe-etal-2018-robust}.

\subsection{Step 1: Embedding Normalization}
\label{sub:normalization}
To directly quote the original paper, the first step of the proposed method is to length normalize the word embeddings $\mathbf{X}_s$ and $\mathbf{X}_t$, then mean center each dimension and finally, length normalize again.

\subsection{Step 2: Fully Unsupervised Initialization}
\label{sub:unsupervised}
The next step is a component introduced by the authors in the original paper: the unsupervised seed dictionary initialization.
To build the initial dictionary $\mathbf{D}^0$, we begin by applying multiple transformations to both $\mathbf{X_s}$ and $\mathbf{X_t}$, described as follows. Given one language's embedding matrix $\mathbf{X}$, we start by computing its Singular Value Decomposition, namely $\mathbf{U}\mathbf{S}\mathbf{V}^{T} = \mathbf{X}$.
We then compute $\sqrt{\mathbf{M}} = \mathbf{U}\mathbf{S}\mathbf{U}^T$ where $\mathbf{M}= \mathbf{X}\mathbf{X}^T = \mathbf{U}\mathbf{S}^2\mathbf{U}^T$ corresponds to the similarity matrix for the given language's embedding matrix.
Each row (word embedding) of the yielded matrices $\sqrt{\mathbf{M}_s}$ and $\sqrt{\mathbf{M}_t}$ is then sorted independently of other rows and we apply the embedding normalization described in \autoref{sub:normalization}
A similarity matrix between the two sets of languages is computed $\mathbf{K} = \sqrt{\mathbf{M}_s}\sqrt{\mathbf{M}_t}^T$.
It is important to note that before the above steps, a vocabulary cutoff of $n=4,000$ is applied, yielding $\mathbf{X}_s, \mathbf{X}_t \in \mathbb{R}^{n \times d}$.

Finally, $\mathbf{D}^0$ is built by applying Cross-domain Similarity Local Scaling (CSLS) \cite{conneau2017word} retrieval on $\mathbf{K}$ and bidirectional dictionary induction $\mathbf{D}^0 = \mathbf{D}_{\mathbf{X}_s \rightarrow \mathbf{X}_t} + \mathbf{D}_{\mathbf{X}_s \leftarrow \mathbf{X}_t}$. Further details on both CSLS and the bidirectional dictionary induction can be found in \autoref{sub:selflearning}


\subsection{Step 3: Robust Self-Learning}
\label{sec:selflearning}

The initial dictionary $\mathbf{D}^0$ is rarely a good solution in itself.
To overcome this, the authors proposed a self-learning algorithm that iteratively refines the previously induced dictionary.
\newcite{Hartmann2019ComparingUW} specifically demonstrated that without this algorithm, the unsupervised dictionary induction is worse than vanilla GAN methods such as \newcite{conneau2017word}.
The algorithm comprises two main steps done iteratively until convergence. The first step is to compute the optimal orthogonal mapping maximizing the objective function \footnote{Here $\mathbf{X}[i, *]$ denotes the $i$-th row of the matrix $\mathbf{X}$}
\begin{equation}
    \underset{\mathbf{W}_s, \mathbf{W}_t}{\argmax} \sum_i \sum_j \mathbf{D}_{ij}\big((\mathbf{X}_s[i, *] \cdot \mathbf{W}_s) \cdot (\mathbf{X}_t[j, *] \cdot \mathbf{W}_t)\big)
    \label{eq:obj_function}
\end{equation}
for the current dictionary $\mathbf{D}^t$ at iteration $t$. The second step is then to apply nearest neighbor retrieval over the similarity matrix of the mapped embeddings $\mathbf{X}_s\mathbf{W}_s\mathbf{W}_t^T\mathbf{X}_t$ to yield the next seed dictionary $\mathbf{D}^{t+1}$ for the next iteration. 

In order to make the self-learning more robust and achieve better performance, the authors also propose four improvements to the above algorithm: stochasticity in the dictionary induction, a frequency-based vocabulary cutoff, usage of the CSLS instead of the nearest neighbor to compute the optimal dictionary and usage of a bidirectional approach in the dictionary induction.

The \textbf{frequency-based vocabulary cutoff} only retains the top $n = 20,000$ most frequent words from both embedding matrices. Done after the unsupervised seed dictionary initialization and before the first self-learning step, the objective of this step is to increase the computing efficiency and reduce the complexity of the optimization problem.
The proposed value of $n$ was given without much explanation, only saying that it was working well in practice.
We chose to further analyze the impact of different values of $n$ in \autoref{subsec:hyperparam_robustness}

The authors proposed a \textbf{stochastic feature} that may be vital for the convergence of the algorithm with some languages.
They randomly keep some elements in the similarity matrix yielded at the end of each iteration with a probability of $p$ while the others are ignored.
This encourages the exploration of the search space and allows the dictionary to greatly vary between two iterations when $p$ is small.
As the algorithm starts to converge, the value of $p$ gradually grows to the maximum value of 1.
The initial value of $p$ is set to $p_0=0.1$ and it is multiplied by $p_\factor=2$ every time the objective function \eqref{eq:obj_function} didn't improve of more than a delta value of $\epsilon = 10^{-6}$ in the last 50 iterations. We also chose to further analyze the impact of the $(p_0, p_{factor})$ combination in \autoref{subsec:hyperparam_robustness}

Typically, to compute the optimal dictionary over the mapped embeddings, we use the nearest neighbor retrieval from the source language into the target language but \newcite{Dinu2014ImprovingZL} showed that this approach suffers from the hubness problem.
This phenomenon, where many points are \textit{universal} neighbours to many other points, is an intrinsic problem of high-dimensional spaces \cite{Radovanovic:2010:HSP:1756006.1953015}.
The authors adopted the \textbf{CSLS} introduced by \newcite{conneau2017word} which specifically tackles this problem.
This approach's idea is to penalize \textit{universal} neighbors by subtracting each word's average cosine similarity to its $k$ nearest neighbors in the other language from the cosine similarity result between words from the target and source languages. Since the value used by the authors is $k=10$, as per \newcite{conneau2017word}'s recommendation, we again chose to further analyze the impact of different values of $k$ in \autoref{subsec:hyperparam_robustness} to grasp a better understanding of its impact.

The authors also proposed to use a \textbf{bidirectional approach for the dictionary induction}.
This improvement is based on the intuition that, when the dictionary is induced from the source into the target language, some of the words may not be present or some may occur numerous times.
The authors claim that those target words occurring multiple times may cause a problem of local optima since they may act as an aggregation hub, making it much more difficult to escape from that undesired solution.
The bidirectional approach thus uses the concatenation of both mappings, source to target and target to source as the dictionary $\mathbf{D}$, namely $\mathbf{D} = \mathbf{D}_{\mathbf{X}_s \rightarrow \mathbf{X}_t} + \mathbf{D}_{\mathbf{X}_s \leftarrow \mathbf{X}_t}$.

\subsection{Step 4: Symmetric Re-Weighting}
\label{sub:reweighting}
As explained in \newcite{artetxe2018aaai}, re-weighting the parameters of the target language's embeddings according to their cross-correlation is beneficial and greatly improved the quality of the induced dictionary.
They also showed that using re-weighting and self-learning didn't seem to work well together since it provokes an accentuation of the local optima problem and discourages the exploration of other possible better regions, which is most of the problem addressed by the four improvements proposed by the authors in the self-learning step.
As a result, this step is done only once after the self-learning loop converged.
However, unlike \newcite{artetxe2018aaai} which applied the re-weighting on either the source or the target language, the authors applied the re-weighting to both languages.
Using the symmetric approach improves the performance of the system, but it's not clear why they chose to use a symmetric re-weighting instead of a target only re-weighting as proposed in \newcite{artetxe2018aaai}.


\section{Reproducing the Results}
\label{sec:reproducing}

We focused on reproducing the results of the entire ablation study of Table 4, as well as the ``\textit{Proposed Method}'' line from Tables 2 and 3 of the original paper of \newcite{artetxe-etal-2018-robust}.
The results comprise four different languages, Deutsch (DE), Spanish (ES), Finnish (FI) and Italian (IT).
Since we did not have access to the dataset of \newcite{zhang-etal-2017-adversarial}, we could not reproduce the results of Table 1.
We thus discuss in this section the results we obtained for the proposed method and the ablation study with some issues we faced.

\subsection{Original Results}
\label{subsec:original_results}

\begin{table*}[h!]
\small
\centering
\setlength{\tabcolsep}{5pt} 
\renewcommand{\arraystretch}{1} 
\begin{tabular}{lcccccccccccccccc}
\toprule
 & \multicolumn{4}{c}{\textbf{EN-DE}} & \multicolumn{4}{c}{\textbf{EN-ES}} & \multicolumn{4}{c}{\textbf{EN-FI}} & \multicolumn{4}{c}{\textbf{EN-IT}}\\
\cmidrule(r{.3em}l){2-5}
\cmidrule(r{.3em}l{.3em}){6-9}
\cmidrule(r{.3em}l{.3em}){10-13}
\cmidrule(rl{.3em}){14-17}
 & best & avg & s & t & best & avg & s & t & best & avg & s & t & best & avg & s & t\\
\cmidrule(rl){1-17}
Original & 48.5 & 48.2 & 1.0 & 7.3 & 37.6 & 37.3 & 1.0 & 9.1 & 33.5 & 32.6 & 1.0 & 12.9 & 48.5 & 48.1 & 1.0 & 8.9\\
\textbf{Reproduced} & 48.5 & 48.3 & 1.0 & \textbf{31.1} & 37.8 & 37.2 & 1.0 & \textbf{35.3} & 33.7 & 32.9 & 1.0 & \textbf{38.1} & 48.5 & 48.2 & 1.0 & \textbf{29.4}\\
\bottomrule
\end{tabular}
\caption{The original results were taken from the original paper of Artetxe et al., (2018). The reproduced results have been generated using their original codebase. We report the best accuracy (best), average accuracy (avg), success rate (s) and average runtime in minutes (t). \textbf{Bold} values represent significant differences between the original and reproduced results.}
\label{table:original_results}
\end{table*}

To reproduce the results of the original paper, we directly used their publicly available codebase\footnote{\url{https://github.com/artetxem/vecmap}}, instead of completely reimplementing their algorithm on our side.
As reported in \autoref{table:original_results}, we were able to reproduce the original results with their codebase within a negligible difference, most likely due to the stochastic nature of the dictionary induction of the algorithm.
Like in the original paper, we provide the best and average (avg) accuracy for every language pair as well as its average runtime (t).
We performed 25 runs per target language and, instead of listing the number of successful runs (where accuracy $> 5~\%$), we present the success \textit{rate} (s).
We can see that, as expected, we have a success rate of 1.0.

The execution time highly differs from the original paper.
It is important to note that even by using the same hardware as the authors (Nvidia Titan Xp GPU), the average runtime for each language is 2 to 4 times longer than the actual runtime reported in the paper.
It is an important factor when comes the time to reproduce the results if we have a limited amount of resources at hand.

Another thing to keep in mind when using this algorithm is that the frequency-based vocabulary cutoff \textbf{assumes that the word vectors have been saved in the embedding text file ordered by their frequency in the training corpus}.
While this is the default behavior of the Fasttext library \cite{mikolov2018advances}, it may not always be the case.


\subsection{Ablation Study}
\label{subsec:ablation_study}

Our reproduction results of the ablation study in the original paper are reported in \autoref{table:ablation_study}. 
Amongst other things, we note that the accuracy results we obtained are all within the 95 \% confidence interval given by our 25 runs, with the only exceptions being the unsupervised initialization ablation and the runtimes. 

Regarding the unsupervised initialization ablation, we initially faced the challenge of having to reproduce the random seed dictionary initialization that was mentioned in the paper yet missing in the code. 
We therefore explored two very plausible approaches to random initializations: one where each word of the smallest language is randomly assigned a word from the biggest language (referred to as `\textit{Random Complete}') and one where a cutoff is done on both languages before the random pairing (referred to as `\textit{Random Cutoff}'). 
The first thing to note is that both our tested random initializations reach convergence between 10 and 30~\% of the time, in contrast with the authors' 0~\% success rate. 
Also, when runs beginning with random initialization are successful, the final performance of the algorithm is the same as the one with the full system. This hints that the initial seed dictionary used, whether obtained by unsupervised or random initialization, only affects the difficulty of the optimization problem but not the retrieved solution.

Our results also showed great differences in the algorithm's runtimes, even though we used the same graphics card as the original paper. 
We also point out that not all ablation study configurations can be run with the same compute power, i.e. when removing the vocabulary cutoff parameter, the matrices no longer fit inside a GPU's memory and we have to prepare additional RAM space and CPU resources to run the script. 
However, when we attempted to reproduce the frequency-based vocabulary cutoff ablation where $n$
is set to 100k, we were unable to obtain a single run to complete even after 3 days of computations. This is why we left this line dashed out in \autoref{table:ablation_study}.

\begin{table*}[h!]
\small
\centering
\setlength{\tabcolsep}{5pt} 
\renewcommand{\arraystretch}{1} 
\begin{tabular}{lcccccccccccccccc}
\toprule
 & \multicolumn{4}{c}{\textbf{EN-DE}} & \multicolumn{4}{c}{\textbf{EN-ES}} & \multicolumn{4}{c}{\textbf{EN-FI}} & \multicolumn{4}{c}{\textbf{EN-IT}}\\
\cmidrule(r{.3em}l){2-5}
\cmidrule(r{.3em}l{.3em}){6-9}
\cmidrule(r{.3em}l{.3em}){10-13}
\cmidrule(rl{.3em}){14-17}
 & best & avg & s & t & best & avg & s & t & best & avg & s & t & best & avg & s & t\\
\cmidrule(rl){1-17}
Full System & 48.5 & 48.2 & 1.0 & 7.3 & 37.6 & 37.3 & 1.0 & 9.1 & 33.5 & 32.6 & 1.0 & 12.9 & 48.5 & 48.1 & 1.0 & 8.9\\
\textbf{Reproduced} & 48.6 & 48.3 & 1.0 & \textbf{35.0} & 37.9 & 37.3 & 1.0 & \textbf{36.2} & 33.8 & 32.9 & 1.0 & \textbf{26.7} & 48.3 & 48.1 & 1.0 & \textbf{30.0}\\
\cmidrule(rl){1-17}
- Unsup. Init. & 0.0 & 0.0 & 0.0 & 17.3 & 0.1 & 0.0 & 0.0 & 15.9 & 0.1 & 0.0 & 0.0 & 13.8 & 0.1 & 0.0 & 0.0 & 16.5\\
\textbf{Rand. Compl.} & \textbf{48.4} & \textbf{14.5} & \textbf{0.3} & \textbf{31.5} & \textbf{37.9} & \textbf{7.6} & \textbf{0.2} & \textbf{20.2} & \textbf{31.7} & \textbf{3.2} & \textbf{0.1} & \textbf{27.6} & \textbf{48.1} & \textbf{9.6} & \textbf{0.2} & \textbf{28.0}\\
\textbf{Rand. Cut.} & \textbf{48.1} & \textbf{14.4} & \textbf{0.3} & \textbf{25.7} & \textbf{38.1} & \textbf{7.6} & \textbf{0.2} & \textbf{24.5} & \textbf{30.0} & \textbf{3.0} & \textbf{0.1} & \textbf{46.1} & \textbf{48.1} & \textbf{19.1} & \textbf{0.4} & \textbf{27.8}\\
\cmidrule(rl){1-17}
- Stochastic & 48.1 & 48.1 & 1.0 & 2.5 & 37.8 & 37.8 & 1.0 & 2.6 & 0.3 & 0.3 & 0.0 & 4.3 & 48.2 & 48.2 & 1.0 & 2.7\\
\textbf{Reproduced} & 48.1 & 48.1 & 1.0 & \textbf{43.0} & 38.1 & 38.1 & 1.0 & \textbf{53.0} & 0.1 & 0.1 & 0.0 & \textbf{26.0} & 48.1 & 48.1 & 1.0 & \textbf{50.0}\\
\cmidrule(rl){1-17}
- Cutoff (n=100k) & 48.3 & 48.1 & 1.0 & 105.3 & 35.5 & 34.9 & 1.0 & 185.2 & 31.9 & 30.8 & 1.0 & 162.5 & 46.9 & 46.5 & 1.0 & 114.5\\
\textbf{Reproduced*} & - & - & - & - & - & - & - & - & - & - & - & - & - & - & - & -\\
\cmidrule(rl){1-17}
- CSLS$^\dagger$ & 0.0 & 0.0 & 0.0 & 13.8 & 0.0 & 0.0 & 0.0 & 14.1 & 0.0 & 0.0 & 0.0 & 13.1 & 0.0 & 0.0 & 0.0 & 15.0\\
\textbf{Reproduced} & 43.1 & 42.8 & 1.0 & 36.9 & 32.9 & 32.7 & 1.0 & 30.2 & 28.0 & 26.9 & 1.0 & 21.9 & 42.9 & 42.5 & 1.0 & 15.0\\
\cmidrule(rl){1-17}
- Bidrectional & 48.3 & 48.0 & 1.0 & 5.5 & 36.2 & 35.8 & 1.0 & 7.3 & 31.4 & 24.9 & 0.8 & 7.8 & 46.0 & 45.4 & 1.0 & 5.6\\
\textbf{Reproduced} & 49.1 & 48.6 & 1.0 & \textbf{36.9} & 37.3 & 37.0 & 1.0 & \textbf{31.2} & 33.1 & 32.0 & 1.0 & \textbf{25.5} & 47.5 & 47.2 & 1.0 & \textbf{31.9}\\
\cmidrule(rl){1-17}
- Re-weighting & 48.1 & 47.4 & 1.0 & 7.0 & 36.0 & 35.5 & 1.0 & 9.1 & 32.9 & 31.8 & 1.0 & 11.2 & 46.1 & 45.6 & 1.0 & 8.4\\
\textbf{Reproduced} & 47.6 & 47.2 & 1.0 & \textbf{35.9} & 37.1 & 36.5 & 1.0 & \textbf{37.0} & 32.0 & 31.5 & 1.0 & \textbf{27.5} & 47.8 & 47.3 & 1.0 & \textbf{31.0}\\
\bottomrule
\end{tabular}
\caption{Ablation study of the algorithm proposed by Artetxe et al., (2018b) on the same four languages. We performed seven ablations and report the original and reproduced best accuracy (best), average accuracy (avg), success rate (s) and average runtime in minutes (t). \textbf{Bold} values represent significant differences between the original and reproduced results. *We did not reproduce this ablation due to time constraints as more than three days per run would have been required to reach convergence. $^\dagger$The authors reported a bug on the code they used for generating the CSLS values which made their code only yield zero accuracies.}
\label{table:ablation_study}
\end{table*}

\subsection{Reproduction Issues}
\label{subsection:reproduction_issues}

While it may seem trivial to use an official implementation to reproduce the results of a paper, the reality is that it often requires a good amount of human effort to run a complete reproduction of the results.
The latter is what happened with us when following the given instructions to obtain the reported results.
While the code did execute and complete when using the \textit{ACL 2018} setting, we had accuracies below the ones expected for each language pair (5 to 7~\% below).
It is only after the further analysis that we found that the provided setting did not include the CSLS procedure, explaining the different results. 
After eventually managing to reproduce the reported full algorithm results, we hit another breaking point: the ablation study was not included in the source code.
While almost all ablations (except the random initialization, as per \autoref{subsec:ablation_study}) could be run from the provided implementation, no script was given to sequentially execute all ablation tests and report the results.
We propose some key recommendations on how to address these issues and facilitate reproducibility and reusability when providing an official implementation with a paper in the \autoref{sec:recommandations}


\section{Recommendations}
\label{sec:recommandations}


Reproducing the model and the results of an original paper can be quite a hassle.
In this section, we provide a general framework applicable to any Machine Learning project that will help researchers deliver highly reproducible experiments. 
We begin with minor recommendations regarding the source code provided by the authors.
We then propose a way to host the dataset and a tool that handles the download and the upload of it.
Since another very important thing to consider when running experiments is to log them all, we hereby propose to automate the logging of the experimentations as well as the gathering of the results.
These steps considerably facilitate the automatic generation of tables and graphs as was required for this challenge.
Finally, we recommend to use a 100~\% reproducible environment to run the experiments, hence to \textit{Dockerize} the whole project \cite{7883438,Hartmann2019ComparingUW}.

\subsection{Codebase Recommendations}

One thing that every research codebase should have is a list of the external libraries needed to execute the code.
In the case of a Python project, it should have a \texttt{requirements.txt} file.
This file holds all the python dependencies needed to run the project's main script.
We thus prepared such a file in our codebase since the original codebase was missing one.

When running experiments, it is important to reduce the number of actions a human needs to perform in order to obtain the final results.
We then made the training and the evaluation of the algorithm, originally in two separate files, into one single script.
This also removes the writing of the mapped embeddings on disk which vastly reduces the amount of disk space needed.

In the same line of ideas, the default hyperparameters of the algorithm should be the ones that reproduce the main results of the paper (Table 2 in \newcite{artetxe-etal-2018-robust}).
This is why we proposed to explicitly \textit{code} not only the full algorithm but also every ablation configuration as \texttt{Experiment} classes within our codebase. 
This abstraction in our source code enables us to easily provide scripts that reproduce our ablation study as well as the hyperparameter grid search conducted in \autoref{subsec:hyperparam_robustness}

Coordination between a paper's key sections and its official implementation is also a concern we wish to raise.
When reading a scientific paper, if one wishes to have a closer look at the implementation of a particular algorithm step or section, one should be able to do so without having to understand the entire codebase.
This is why we created an exact correspondence between step names in the original paper like \textit{'CSLS Retrieval' and 'Robust Self-Learning'} and function names in our source code. 
We argue this name mapping should be easy to implement at the end of the delivery of a research project and that it contributes significantly towards easier reusability of the delivered implementation.

\subsection{Dataset Handling}

Properly handling the benchmark dataset is often an underestimated point. 
In an iterative and collaborative setting, it is important to efficiently host (when possible) \emph{and} version the data. We thus recommend a tool designed to handle those two elements flawlessly; Data Version Control (DVC). 
Similar to standard Version Control Systems (VCS) like Git\footnote{\url{https://git-scm.com/}}, DVC tracks the different state of the dataset during development as well as in between the processing steps before obtaining the final results of the model.

While there is no need in our particular reproducibility challenge context to track the different states of the dataset over time, it definitely requires an efficient collaboration environment, hence our choice to use the Python DVC library \footnote{\url{https://dvc.org}} with Amazon S3 as the remote repository. 
DVC was designed with large data files in mind, meaning gigabytes or even hundreds of gigabytes in file size. 
In our case, the original dataset takes up to 6 Gigabytes. 
The previous way of retrieving the dataset over the network with a standard 20 Mbits/sec internet connexion took up to an hour to complete (including uncompressing the data).
Using DVC reduced the retrieval time of the dataset to 3 minutes over the network with the same internet connexion. While retrieving the dataset may seem like a one-time effort during the development of the model, when comes the time to distribute the computation over several machines, one can save valuable time.
We also made the dataset available as a public archive\footnote{\url{https://vecmap-submission.s3.amazonaws.com/dataset.tar.gz}} since it was required by the challenge.

\subsection{Automatic Experiment Logging}

When doing research, it is easy to enter the experiment's hurry loop; as soon as we have an idea, we code it and we launch our main script without committing the modifications. 
Grossly keeping track of an architecture and its corresponding results in our head or a spreadsheet is good for nothing when it comes to the time to retrieve and analyze past experiments.

We thus propose to automate the process of logging as well as retrieving the results of every experiment in order to reduce the risk of losing experiment information.
To this end, we used a flexible yet emergent tool that beautifully solves this problem; MLflow\footnote{\url{https://mlflow.org}}$^,$\footnote{One can find numerous alternatives such as \href{https://sacred.readthedocs.io/en/stable/}{Sacred}, \href{https://www.comet.ml/}{Comet.ml} or \href{https://www.wandb.com/}{Weights and Biases} for example.}.
MLflow provides a model agnostic Python API that lets you track not only the results of a given configuration but also the associated source code, the dataset used, and much more.
It has been of great use for the automatic generation of tables and graphs in this actual paper as it is required by the challenge. 
We also believe it is vital to use such a framework for any scientific team doing serious research to reduce the overhead and stress of manually logging and keeping the information about experimentations, especially considering the low effort it requires to setup.

\subsection{Dockerization}

Docker\footnote{\url{https://www.docker.com/}} is a software that provides an abstraction of the system libraries, tools, and runtime.
A Docker container is essentially a lightweight executable package that can run on every\footnote{As long as the environment provides the necessary hardware specifications.} environment.
In this project, we did face a dependencies problem between the \texttt{Cupy} python library and its associated CUDA drivers.
In fact, even with a Docker container and the \texttt{nvidia-docker}\footnote{\url{https://github.com/NVIDIA/nvidia-docker}} library, we had to make sure that the \texttt{Cupy} compiled library matched the actual host's CUDA drivers.
This issue brings the reproducibility of the project at stake when the hardware of the host's machine differs from the original one.
We thus assume that the host machine running our codebase within our provided docker image\footnote{registry.gitlab.com/nicolasgarneau/vecmap} has the requirements such as the CUDA drivers to fulfill the experiments.

\section{Assessing the Algorithm's Robustness}
\label{sec:assess}

\begin{table*}[h!]
\small
\centering
\setlength{\tabcolsep}{5pt} 
\renewcommand{\arraystretch}{1} 
\begin{tabular}{lcccccccccccccccc}
\toprule
 & \multicolumn{4}{c}{\textbf{EN-ET}} & \multicolumn{4}{c}{\textbf{EN-FA}} & \multicolumn{4}{c}{\textbf{EN-LV}} & \multicolumn{4}{c}{\textbf{EN-VI}}\\
\cmidrule(r{.3em}l){2-5}
\cmidrule(r{.3em}l{.3em}){6-9}
\cmidrule(r{.3em}l{.3em}){10-13}
\cmidrule(rl{.3em}){14-17}
 & best & avg & s & t & best & avg & s & t & best & avg & s & t & best & avg & s & t\\
\cmidrule(rl){1-17}
\textbf{Vecmap} & 28.5 & 26.7 & 1.0 & 20.5 & 35.6 & 34.5 & 1.0 & 28.7 & 23.4 & \textbf{2.6} & \textbf{0.1} & 41.5 & \textbf{0.0} & \textbf{0.0} & \textbf{0.0} & 35.8\\
\textbf{- Unsup. (Rand.)} & \textbf{0.1} & \textbf{0.0} & \textbf{0.0} & 33.0 & \textbf{0.1} & \textbf{0.0} & \textbf{0.0} & 19.5 & \textbf{0.1} & \textbf{0.0} & \textbf{0.0} & 29.6 & \textbf{0.1} & \textbf{0.0} & \textbf{0.0} & 34.7\\
\textbf{- Unsup. (Rand. Cut.)} & \textbf{0.1} & \textbf{0.0} & \textbf{0.0} & 35.4 & \textbf{0.3} & \textbf{0.0} & \textbf{0.0} & 31.5 & \textbf{0.1} & \textbf{0.0} & \textbf{0.0} & 27.7 & \textbf{0.0} & \textbf{0.0} & \textbf{0.0} & 33.9\\
\textbf{- Stochastic} & \textbf{0.1} & \textbf{0.1} & \textbf{0.0} & 56.3 & \textbf{0.1} & \textbf{0.1} & \textbf{0.0} & 20.0 & \textbf{0.0} & \textbf{0.0} & \textbf{0.0} & 32.0 & \textbf{0.0} & \textbf{0.0} & \textbf{0.0} & 28.0\\
\textbf{- CSLS} & 19.7 & \textbf{2.0} & \textbf{0.1} & 30.5 & 28.0 & 27.5 & 1.0 & 43.7 & \textbf{0.1} & \textbf{0.0} & \textbf{0.0} & 24.1 & \textbf{0.0} & \textbf{0.0} & \textbf{0.0} & 33.2\\
\textbf{- Bidirectional} & 28.1 & 26.9 & 1.0 & 31.7 & 34.7 & 33.9 & 1.0 & 28.7 & \textbf{0.3} & \textbf{0.1} & \textbf{0.0} & 28.8 & \textbf{0.1} & \textbf{0.0} & \textbf{0.0} & 32.5\\
\textbf{- Reweighting} & 28.5 & 26.3 & 1.0 & 22.1 & 34.9 & 33.8 & 1.0 & 30.5 & 23.0 & \textbf{2.5} & \textbf{0.1} & 27.9 & \textbf{0.0} & \textbf{0.0} & \textbf{0.0} & 36.4\\
\bottomrule
\end{tabular}
\caption{Best accuracy (best), average accuracy (avg), success rate (s) and average runtime in minutes (t) of the full system on four other languages, Estonian (ET), Persian (FA), Latvian (LV) and Vietnamese (VI). These languages have been carefully selected to illustrate the robustness of the algorithm. They have very different roots then the English language. We also conducted an ablation study to illustrate the stability of the algorithm on such languages.}
\label{table:other_languages_results}
\end{table*}

\newcite{vulic-etal-2019-really} showed that completely unsupervised word translation approaches tend to fail when language pairs are distant.
They however identify \newcite{artetxe-etal-2018-robust}'s algorithm as the current most robust among completely unsupervised approaches.
Hence, in order to assess ourselves the algorithm's robustness, we decided to apply it on other languages that have fewer similarities with the English language.
We also conduct a grid search over key hyperparameters which enlightens us on the stability of the whole procedure.

\subsection{More Languages}

We carefully selected four new languages that are characterized by very different roots than the one used in the original paper.
We used Estonian (ET) which is a language that gets its root from Finno-Ugric, the same as Finnish.
We also selected Persian (FA), Latvian (LV) and Vietnamese (VI).
We can see in \autoref{table:other_languages_results} that the results on Estonian corroborate the results from the initial paper where the stochastic dictionary induction step is crucial for proper convergence.
We observed similar behavior for the Persian language.
Interestingly, even with the full system, the algorithm poorly performs on Latvian and Vietnamese languages.
We conducted the same ablation study as with the original languages.
We can see that the algorithm does not converge without an unsupervised initialization and without the stochastic procedure.
It also struggles to converge on three languages out of four when the CSLS component is turned off.
These results clearly show that the proposed method may become brittle when the target language shares fewer commonalities with the source language.

\subsection{Robustness to Hyperparameters}
\label{subsec:hyperparam_robustness}

In order to correctly assess the algorithm's robustness to variations in one key hyperparameter's values, we conducted experiments where we fixed all of the parameter values to the default ones and only varied the tested hyperparameter, ensuring adequate conclusions could be drawn. The key parameters we chose to examine are (1) the number of considered neighbors in the CSLS procedure, (2) the number of retained words in the frequency-based vocabulary cutoff and (3) the initial value of $p$ and its growing factor in the stochastic dictionary induction. We then assess each hyperparameter's impact on both the performance and the execution time (in terms of the number of iterations and/or iteration duration) of the algorithm in order to provide well-informed recommendations.

\subsubsection*{CSLS}

\begin{figure*}[th!]
\centering
\begin{tikzpicture}
\begin{axis}[grid style={dashed,gray!50}, axis y line*=left, axis x line*=bottom, every axis plot/.append style={line width=1.25pt, mark size=0pt}, width=.27\textwidth, height=.25\textwidth, grid=major, name=plot0, xshift=-.1\textwidth, y tick label style={/pgf/number format/fixed zerofill, /pgf/number format/precision=1}, xmin=1.0, xmax=20.0, ymin=47.378979514280886, ymax=49.415247120904745, title=English-Deutsch]
\addplot[blue, ylabel near ticks, forget plot, mark=*, line width=1.2pt, mark size=.9pt] table[x=x0, y=y0, col sep=comma]{./grid_search_data/csls_en_de.csv};
\addplot[forget plot, name path=upper, draw=none] table[x=x1, y=y1, col sep=comma]{./grid_search_data/csls_en_de.csv};
\addplot[forget plot, name path=lower, draw=none] table[x=x2, y=y2, col sep=comma]{./grid_search_data/csls_en_de.csv};
\addplot[fill=blue!10] fill between[of=upper and lower];
\end{axis}
\begin{axis}[grid style={dashed,gray!50}, axis y line*=left, axis x line*=bottom, every axis plot/.append style={line width=1.25pt, mark size=0pt}, width=.28\textwidth, height=.25\textwidth, grid=major, at=(plot0.south east), anchor=south west, xshift=-.035\textwidth, name=plot1, y tick label style={/pgf/number format/fixed zerofill, /pgf/number format/precision=1}, xmin=1.0, xmax=20.0, ymin=35.34151463811338, ymax=38.44547310349182, title=English-Spanish]
\addplot[blue, ylabel near ticks, forget plot, mark=*, line width=1.2pt, mark size=.9pt] table[x=x3, y=y3, col sep=comma]{./grid_search_data/csls_en_es.csv};
\addplot[forget plot, name path=upper, draw=none] table[x=x4, y=y4, col sep=comma]{./grid_search_data/csls_en_es.csv};
\addplot[forget plot, name path=lower, draw=none] table[x=x5, y=y5, col sep=comma]{./grid_search_data/csls_en_es.csv};
\addplot[fill=blue!10] fill between[of=upper and lower];
\end{axis}
\begin{axis}[grid style={dashed,gray!50}, axis y line*=left, axis x line*=bottom, every axis plot/.append style={line width=1.25pt, mark size=0pt}, width=.28\textwidth, height=.25\textwidth, grid=major, at=(plot1.south east), anchor=south west, xshift=0.03\textwidth, name=plot2, y tick label style={/pgf/number format/fixed zerofill, /pgf/number format/precision=1}, xmin=1.0, xmax=20.0, ymin=29.95888181700087, ymax=34.83680711898751, title=English-Finnish]
\addplot[blue, ylabel near ticks, forget plot, mark=*, line width=1.2pt, mark size=.9pt] table[x=x6, y=y6, col sep=comma]{./grid_search_data/csls_en_fi.csv};
\addplot[forget plot, name path=upper, draw=none] table[x=x7, y=y7, col sep=comma]{./grid_search_data/csls_en_fi.csv};
\addplot[forget plot, name path=lower, draw=none] table[x=x8, y=y8, col sep=comma]{./grid_search_data/csls_en_fi.csv};
\addplot[fill=blue!10] fill between[of=upper and lower];
\end{axis}
\begin{axis}[grid style={dashed,gray!50}, axis y line*=left, axis x line*=bottom, every axis plot/.append style={line width=1.25pt, mark size=0pt}, width=.28\textwidth, height=.25\textwidth, grid=major, at=(plot2.south east), anchor=south west, xshift=.09\textwidth, name=plot3, y tick label style={/pgf/number format/fixed zerofill, /pgf/number format/precision=1}, xmin=1.0, xmax=20.0, ymin=46.049928359832506, ymax=49.39452462425433, title=English-Italian]
\addplot[blue, ylabel near ticks, forget plot, mark=*, line width=1.2pt, mark size=.9pt] table[x=x9, y=y9, col sep=comma]{./grid_search_data/csls_en_it.csv};
\addplot[forget plot, name path=upper, draw=none] table[x=x10, y=y10, col sep=comma]{./grid_search_data/csls_en_it.csv};
\addplot[forget plot, name path=lower, draw=none] table[x=x11, y=y11, col sep=comma]{./grid_search_data/csls_en_it.csv};
\addplot[fill=blue!10] fill between[of=upper and lower];
\end{axis}
\end{tikzpicture}
\caption{Average accuracy percentage results per number of considered neighbors in the \textbf{CSLS} procedure on the various language pairs. All reported results are obtained after a total of 10 runs per value and the shaded region represents a 95 \% confidence interval on the accuracy mean.}
\label{figure:csls}
\end{figure*}
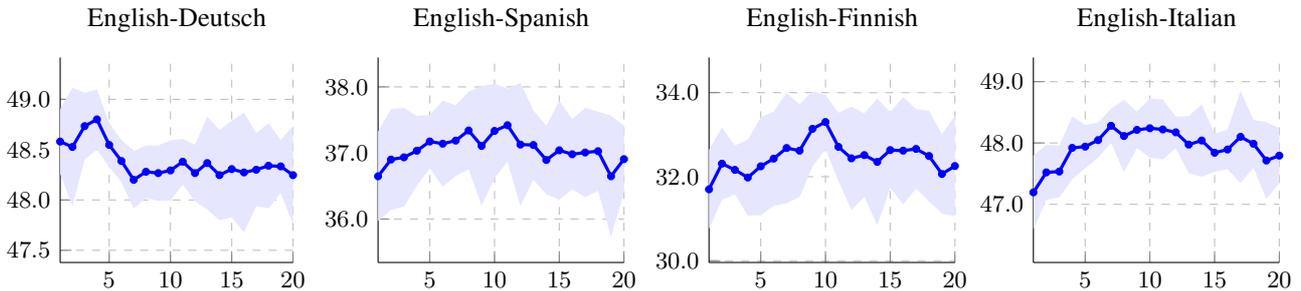

We conducted experiments where we varied the $k$ number of neighbors considered in the CSLS procedure from 1 to 20, with results reported on \autoref{figure:csls}. For all language pairs, we denote a variation of approximately 1~\% between the highest and lowest accuracy obtained over the evaluated range. These variations are however well in between the 95~\% confidence interval region for most of the tested values, suggesting the correlation between the performance and the number of neighbors considered in CSLS is loose. Furthermore, when taking into account that iteration duration only slightly grows with the growth of $k$, the author's suggested universal value of $k=10$ neighbors considered in the CSLS retrieval procedure appears like a legitimate compromise.

\subsubsection*{Frequency-based vocabulary cutoff}

For this experiment, we only retained the $n$ most frequent words of both languages before launching the  self-learning iterative procedure (\autoref{sec:selflearning}), with values of $n$ ranging from 10k to 30k, with increments of 1k.
When increasing the value of $n$, our results show the system's accuracy decreasing on Spanish, increasing on both Finnish and Deutsch and attaining a stable range for Italian.
While the accuracy difference is very different from one target language to another, the variation on all the tested range is between 1 and 2~\%.
Regarding the iteration duration's correlation with the number of retained words, our experiments show a quadratic growth of an iteration's duration as well as an overall increase in the number of iterations before convergence when increasing $n$, in line with the original paper's conclusion.
Pairing the highly language-dependent impact of this hyperparameter on the algorithm's performance with its major impact on its execution time, we conclude that the number of most frequent words retained before the self-learning procedure should be the target of careful finetuning for each target language.

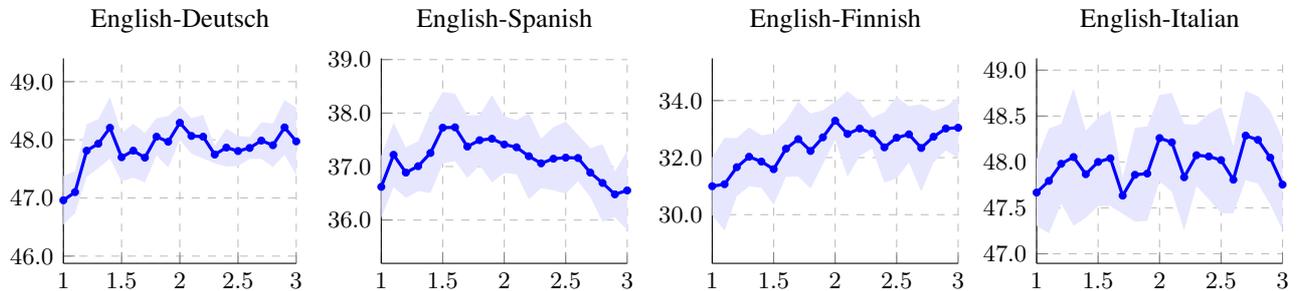
\begin{figure*}[th!]
\centering
\begin{tikzpicture}
\begin{axis}[grid style={dashed,gray!50}, axis y line*=left, axis x line*=bottom, every axis plot/.append style={line width=1.25pt, mark size=0pt}, width=.27\textwidth, height=.25\textwidth, grid=major, name=plot0, xshift=-.1\textwidth, y tick label style={/pgf/number format/fixed zerofill, /pgf/number format/precision=1}, xmin=1.0, xmax=3.0, ymin=45.87520000000001, ymax=49.400715395978615, title=English-Deutsch]
\addplot[blue, ylabel near ticks, forget plot, mark=*, line width=1.2pt, mark size=.9pt] table[x=x12, y=y12, col sep=comma]{./grid_search_data/voc_cutoff_en_de.csv};
\addplot[forget plot, name path=upper, draw=none] table[x=x13, y=y13, col sep=comma]{./grid_search_data/voc_cutoff_en_de.csv};
\addplot[forget plot, name path=lower, draw=none] table[x=x14, y=y14, col sep=comma]{./grid_search_data/voc_cutoff_en_de.csv};
\addplot[fill=blue!10] fill between[of=upper and lower];
\end{axis}
\begin{axis}[grid style={dashed,gray!50}, axis y line*=left, axis x line*=bottom, every axis plot/.append style={line width=1.25pt, mark size=0pt}, width=.28\textwidth, height=.25\textwidth, grid=major, at=(plot0.south east), anchor=south west, xshift=-.035\textwidth, name=plot1, y tick label style={/pgf/number format/fixed zerofill, /pgf/number format/precision=1}, xmin=1.0, xmax=3.0, ymin=35.18970203311149, ymax=39.02331065964951, title=English-Spanish]
\addplot[blue, ylabel near ticks, forget plot, mark=*, line width=1.2pt, mark size=.9pt] table[x=x15, y=y15, col sep=comma]{./grid_search_data/voc_cutoff_en_es.csv};
\addplot[forget plot, name path=upper, draw=none] table[x=x16, y=y16, col sep=comma]{./grid_search_data/voc_cutoff_en_es.csv};
\addplot[forget plot, name path=lower, draw=none] table[x=x17, y=y17, col sep=comma]{./grid_search_data/voc_cutoff_en_es.csv};
\addplot[fill=blue!10] fill between[of=upper and lower];
\end{axis}
\begin{axis}[grid style={dashed,gray!50}, axis y line*=left, axis x line*=bottom, every axis plot/.append style={line width=1.25pt, mark size=0pt}, width=.28\textwidth, height=.25\textwidth, grid=major, at=(plot1.south east), anchor=south west, xshift=0.03\textwidth, name=plot2, y tick label style={/pgf/number format/fixed zerofill, /pgf/number format/precision=1}, xmin=1.0, xmax=3.0, ymin=28.292753053731687, ymax=35.48002577918472, title=English-Finnish]
\addplot[blue, ylabel near ticks, forget plot, mark=*, line width=1.2pt, mark size=.9pt] table[x=x18, y=y18, col sep=comma]{./grid_search_data/voc_cutoff_en_fi.csv};
\addplot[forget plot, name path=upper, draw=none] table[x=x19, y=y19, col sep=comma]{./grid_search_data/voc_cutoff_en_fi.csv};
\addplot[forget plot, name path=lower, draw=none] table[x=x20, y=y20, col sep=comma]{./grid_search_data/voc_cutoff_en_fi.csv};
\addplot[fill=blue!10] fill between[of=upper and lower];
\end{axis}
\begin{axis}[grid style={dashed,gray!50}, axis y line*=left, axis x line*=bottom, every axis plot/.append style={line width=1.25pt, mark size=0pt}, width=.28\textwidth, height=.25\textwidth, grid=major, at=(plot2.south east), anchor=south west, xshift=.09\textwidth, name=plot3, y tick label style={/pgf/number format/fixed zerofill, /pgf/number format/precision=1}, xmin=1.0, xmax=3.0, ymin=46.89575650054789, ymax=49.12788833391089, title=English-Italian]
\addplot[blue, ylabel near ticks, forget plot, mark=*, line width=1.2pt, mark size=.9pt] table[x=x21, y=y21, col sep=comma]{./grid_search_data/voc_cutoff_en_it.csv};
\addplot[forget plot, name path=upper, draw=none] table[x=x22, y=y22, col sep=comma]{./grid_search_data/voc_cutoff_en_it.csv};
\addplot[forget plot, name path=lower, draw=none] table[x=x23, y=y23, col sep=comma]{./grid_search_data/voc_cutoff_en_it.csv};
\addplot[fill=blue!10] fill between[of=upper and lower];
\end{axis}
\end{tikzpicture}
\caption{Average accuracy percentage results per number of retained words (in tens of thousands) in the \textbf{frequency-based vocabulary cutoff} method on the various language pairs. All reported results are obtained after a total of 10 runs per value and the shaded region represents a 95 \% confidence interval on the accuracy mean.}
\label{figure:voc_cutoff}
\end{figure*}

\subsubsection*{Stochastic dictionary induction}

For the tests on the stochastic dictionary induction, we considered a linear space of 5 values between 0.05 and 0.3 for the initial keep probability ($p_0$) and a linear space of 4 values between 1.5 and 3 for $p$'s growth factor ($p_\factor$) and ran tests for each of the 20 total combinations. Our results only show a slight performance difference between all tested value pairs, with all language pairs only varying for less than 1~\% and three of the four language pairs varying for less than 0.5~\%. One important to note however is that making the algorithm greedier (with a higher value of $p_0$) does not lead to any performance loss: the best performances are rather found when using those high $p_0$ values. Considering the number of iterations only decreases when $p_0$ grows, it appears the original paper's $p_0=0.1$ value only increases the number of iterations without a significant impact on performance. No such conclusion can be drawn for the $p_\factor$ hyperparameter, which appears very weakly correlated to overall performance.

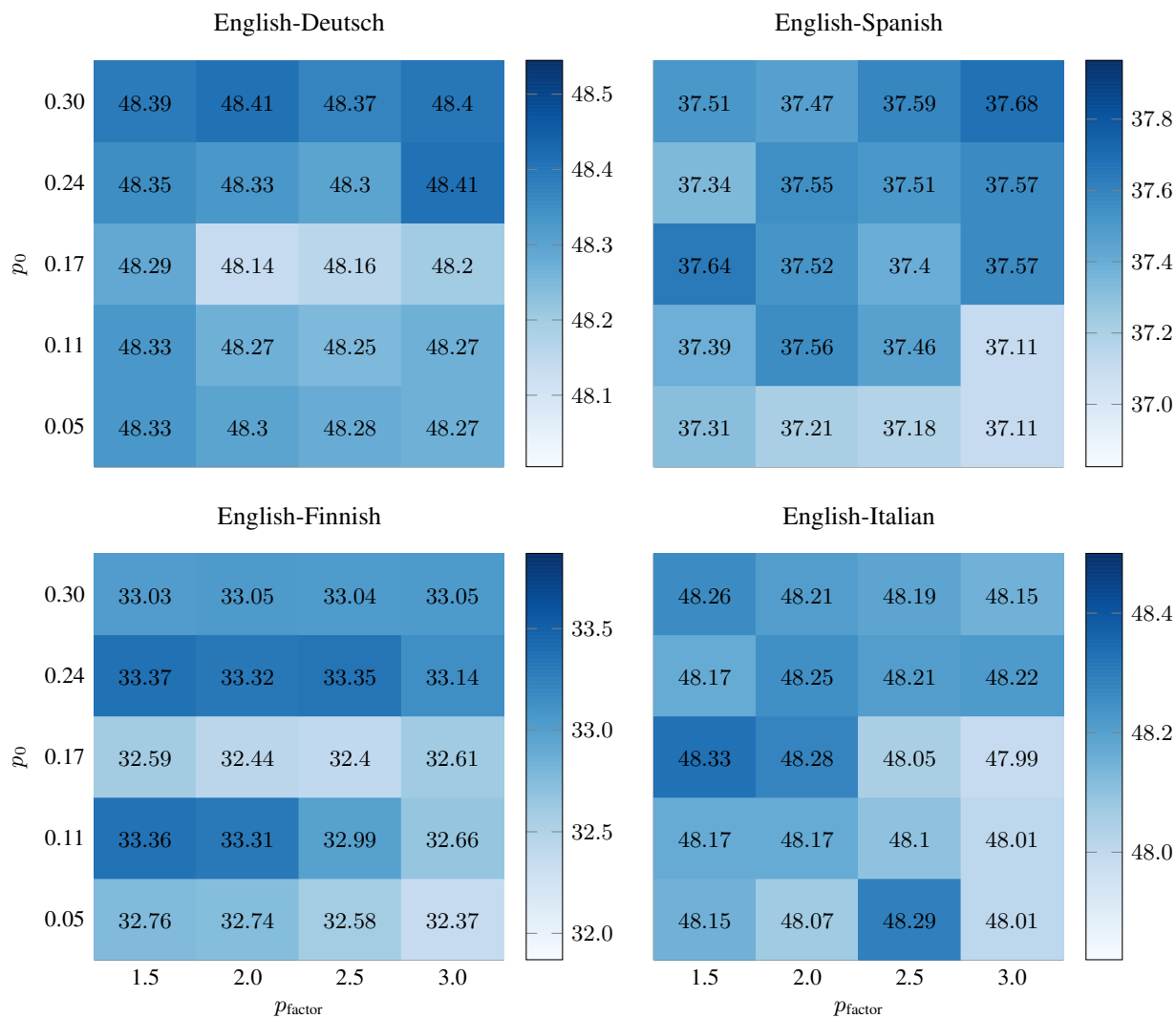
\begin{figure*}[th!]
\centering
\begin{tikzpicture}
\begin{axis}[grid style={dashed,gray!50}, axis y line*=left, axis x line*=bottom, colorbar, nodes near coords={\pgfmathprintnumber\pgfplotspointmeta}, every node near coord/.append style={xshift=0pt,yshift=-7pt, black, font=\footnotesize}, every axis plot/.append style={line width=0pt, mark size=0pt}, width=.42\textwidth, height=.42\textwidth, grid=none, enlargelimits=false, name=plot0, xticklabels={\empty}, ylabel=$p_0$, yticklabels={0.05,0.11,0.17,0.24,0.30}, xtick={0.000,1.000,2.000,3.000}, ytick={0.000,1.000,2.000,3.000,4.000}, colorbar style={/pgf/number format/fixed zerofill, /pgf/number format/precision=1}, title=English-Deutsch]
\addplot[matrix plot*, point meta min=48.005, point meta max=48.544999999999995, point meta=explicit, mesh/rows=5, mesh/cols=4] table[x=x24, y=y24, meta=z24, col sep=comma]{./grid_search_data/stochastic_en_de.csv};
\end{axis}
\begin{axis}[grid style={dashed,gray!50}, axis y line*=left, axis x line*=bottom, colorbar, nodes near coords={\pgfmathprintnumber\pgfplotspointmeta}, every node near coord/.append style={xshift=0pt,yshift=-7pt, black, font=\footnotesize}, every axis plot/.append style={line width=0pt, mark size=0pt}, width=.42\textwidth, height=.42\textwidth, grid=none, enlargelimits=false, at=(plot0.south east), anchor=south west, xshift=.12\textwidth, name=plot1, xticklabels={\empty}, yticklabels={\empty}, xtick={0.000,1.000,2.000,3.000}, ytick={0.000,1.000,2.000,3.000,4.000}, colorbar style={/pgf/number format/fixed zerofill, /pgf/number format/precision=1}, title=English-Spanish]
\addplot[matrix plot*, point meta min=36.825, point meta max=37.965, point meta=explicit, mesh/rows=5, mesh/cols=4] table[x=x25, y=y25, meta=z25, col sep=comma]{./grid_search_data/stochastic_en_es.csv};
\end{axis}
\begin{axis}[grid style={dashed,gray!50}, axis y line*=left, axis x line*=bottom, colorbar, nodes near coords={\pgfmathprintnumber\pgfplotspointmeta}, every node near coord/.append style={xshift=0pt,yshift=-7pt, black, font=\footnotesize}, every axis plot/.append style={line width=0pt, mark size=0pt}, width=.42\textwidth, height=.42\textwidth, grid=none, enlargelimits=false, at=(plot0.south west), anchor=north west, yshift=-0.07\textwidth, name=plot2, xlabel=$p_\factor$, xticklabels={1.5,2.0,2.5,3.0}, ylabel=$p_0$, yticklabels={0.05,0.11,0.17,0.24,0.30}, xtick={0.000,1.000,2.000,3.000}, ytick={0.000,1.000,2.000,3.000,4.000}, colorbar style={/pgf/number format/fixed zerofill, /pgf/number format/precision=1}, title=English-Finnish]
\addplot[matrix plot*, point meta min=31.869999999999997, point meta max=33.87, point meta=explicit, mesh/rows=5, mesh/cols=4] table[x=x26, y=y26, meta=z26, col sep=comma]{./grid_search_data/stochastic_en_fi.csv};
\end{axis}
\begin{axis}[grid style={dashed,gray!50}, axis y line*=left, axis x line*=bottom, colorbar, nodes near coords={\pgfmathprintnumber\pgfplotspointmeta}, every node near coord/.append style={xshift=0pt,yshift=-7pt, black, font=\footnotesize}, every axis plot/.append style={line width=0pt, mark size=0pt}, width=.42\textwidth, height=.42\textwidth, grid=none, enlargelimits=false, at=(plot2.south east), anchor=south west, xshift=.12\textwidth, name=plot3, xlabel=$p_\factor$, xticklabels={1.5,2.0,2.5,3.0}, yticklabels={\empty}, xtick={0.000,1.000,2.000,3.000}, ytick={0.000,1.000,2.000,3.000,4.000}, colorbar style={/pgf/number format/fixed zerofill, /pgf/number format/precision=1}, title=English-Italian]
\addplot[matrix plot*, point meta min=47.82000000000001, point meta max=48.5, point meta=explicit, mesh/rows=5, mesh/cols=4] table[x=x27, y=y27, meta=z27, col sep=comma]{./grid_search_data/stochastic_en_it.csv};
\end{axis}
\end{tikzpicture}
\caption{Average accuracy percentage results per initial value of $p$ ($p_0$) and its growing factor ($p_\factor$) in the \textbf{stochastic dictionary induction} on the various language pairs. All reported results are obtained after a total of 10 runs per ($p_0$, $p_\factor$) pair.}
\label{figure:stochastic}
\end{figure*}

\section{Conclusion}
\label{sec:conclusion}

In this paper, we studied the reproducibility of the model proposed by \newcite{artetxe-etal-2018-robust}.
We found out that their method of mapping embeddings between two languages is robust when the languages share commonalities.
Otherwise, the approach struggles to learn proper mapping.
We also assessed the robustness of the hyperparameters of the algorithm in many languages.

We introduced several recommendations regarding the guidelines every researcher should follow in order to deliver fully reproducible research.
It is often said that replicability (reproducing the results of a model from a new implementation) is more complicated than reproducibility (reproducing the results from an existing implementation).
However, we found out that reproducing the results may become an issue if there are hardware or time constraints as we faced during our experimentations.
Indeed, we were able to perform a grid-search on the hyperparameter and validate the robustness of the algorithm thanks to the 64 GPUs we had in hand, otherwise, it would have taken months to run.
That being said, reproducibility is an issue when hardware and time constraints come into play.

\section{Acknowledgements}
This research was enabled in part by support provided by Calcul Qu\'ebec (https://www.calculquebec.ca/) and Compute Canada (www.computecanada.ca). We also acknowledge the support of the Natural Sciences and Engineering Research Council of Canada (NSERC).
Finally, we wish to thank Anders S\o gaard for his precious advice and the reviewers for their insightful comments regarding our work and methodology.

\section{References}
\label{main:ref}

\bibliographystyle{lrec}
\bibliography{lrec2020-REPROLANG}

\end{document}